\documentclass[runningheads]{llncs}
\usepackage{graphicx}
\usepackage{tikz}
\usepackage{comment}
\usepackage{amsmath,amssymb} % define this before the line numbering.
\usepackage{color}

\usepackage[accsupp]{axessibility}  % Improves PDF readability for those with disabilities.

\begin{document}
% \linenumbers
\pagestyle{headings}
\mainmatter
\def\ECCVSubNumber{2571}  % Insert your submission number here

\title{End-to-End Learning of Multi-category 3D Pose and Shape Estimation} % Replace with your title

%\begin{comment}
% \titlerunning{ECCV-22 submission ID \ECCVSubNumber} 
% \authorrunning{ECCV-22 submission ID \ECCVSubNumber} 
% \author{Anonymous ECCV submission}
% \institute{Paper ID \ECCVSubNumber}
%\end{comment}
%******************

% \begin{comment}
\author{Yigit Baran Can\inst{1} \and
Alexander Liniger\inst{1} \and
Danda Pani Paudel\inst{1}\and
Luc Van Gool\inst{1,2}}
\authorrunning{Can et al.}
\institute{Computer Vision Lab, ETH Zurich \and
VISICS, ESAT/PSI, KU Leuven
\email{lncs@springer.com}\\
\email{\{yigit.can, alex.liniger, paudel, vangool\}@vision.ee.ethz.ch}}

% \author{ Yigit Baran Can\textsuperscript{1}\space\space\space\space Alexander Liniger\textsuperscript{1}\space\space\space\space Danda Pani Paudel\textsuperscript{1}\space\space\space\space Luc Van Gool\textsuperscript{1,2}\\
% \textsuperscript{1}Computer Vision Lab, ETH Zurich\space\space\space\space \textsuperscript{2}VISICS, ESAT/PSI, KU Leuven \\ {\tt\small $\{$yigit.can, alex.liniger, paudel, vangool$\}$@vision.ee.ethz.ch} }

% \end{comment}
%******************
\maketitle
\begin{abstract}

In this paper, we study the representation of the shape and pose of objects using their keypoints. Therefore, we propose an end-to-end method that simultaneously detects 2D keypoints from an image and lifts them to 3D. The proposed method learns both 2D detection and 3D lifting only from 2D keypoints annotations. In addition to being end-to-end from images to 3D keypoints, our method also  handles objects from multiple categories using a single neural network. We use a Transformer-based architecture to detect the keypoints, as well as to summarize the visual context of the image. This visual context information is then used while lifting the keypoints to 3D, to allow context-based reasoning for better performance. Our method can handle occlusions as well as a wide variety of object classes. Our experiments on three benchmarks demonstrate that our method performs better than the state-of-the-art. Our source code will be made publicly available.  

\end{abstract}

%%%%%%%%% BODY TEXT
\section{Introduction}
\label{sec:intro}

% Keypoint-based shape and pose representations, often in some wire-frame form connecting human body and hand joints, are attractive because of their simplicity and ease of handling. 
A keypoint-based shape and pose representation is attractive because of its simplicity and ease of handling. 
Example applications include 3D reconstruction \cite{novotny2019c3dpo,DBLP:journals/ijcv/DaiLH14,Snavely2007}, registration \cite{yew20183dfeat,Kneip2014,Luong1995,Loper2015}, and human body pose analysis~\cite{shotton2011real,moreno20173d,cao2017,bogo2016smpl}, recognition \cite{he2017mask,sattler2011fast}, and generation \cite{tang2019cycle,zafeiriou20173d}. The keypoints are often detected as 2D image coordinates due to the ease of the corresponding annotation. Yet, this often does not suffice for the subsequent geometric reasoning tasks. In many applications (e.g. augmented reality), both 3D shape and pose are required~\cite{tulsiani2015viewpoints}.

Therefore, it stands to reason that keypoints should be estimated in 3D, similar to~\cite{zhao2020learning,suwajanakorn2018discovery,tulsiani2015viewpoints,sundermeyer2020augmented}. Such solutions, however, come with one or both of the two pitfalls: (i) need of 3D keypoints, pose, or multiple views for supervision; (ii) the lack of direct pose reasoning with respect to a canonical frame.  
In this regard, learning-based methods can provide both 3D keypoints and pose merely from a single image, making them suitable for a very wide range of applications from scene understanding~\cite{fernandez2020indoor} to augmented reality~\cite{marchand2015pose}.
Alternatively, template-based single view methods~\cite{yang2020perfect,shi2021optimal} may also be used to obtain 3D keypoints and pose from 2D keypoints. However, template-based methods, besides requiring templates, are known to be sensitive to self-occlusions~\cite{dang20203d}. Therefore, we adopt a learning-based method for single view inference of both the 3D keypoints and the pose of objects. 

In this paper, we consider that only one image per object is available both during training and inference. This assumption allows us to learn from diverse datasets, such as internet image collections, potentially offering us a high generalization ability. For better scalability, we also assume that only minimalistic supervision in the form of 2D keypoints and objects' categories are available. In essence, we wish to learn the 3D keypoints and pose of objects from an image collection of multiple categories, where not even multi-view images of the same object are available. 

Existing methods that learn 3D shape and pose from an image collection by object categories are also known as deep non-rigid structure-from-motion (NrSfM) due to their underlying assumption. The method proposed in this paper also belongs to the same class, which can be divided into single~\cite{DBLP:conf/eccv/ParkLK20,DBLP:conf/iccv/KongL19,wang2021paul,zeng2021pr} and multi-category~\cite{novotny2019c3dpo} methods. Multi-category methods are strikingly more interesting due to two main reasons, (i) computational: one single neural network can infer shapes and poses for objects from different categories; (ii) relational: possibility of establishing/exploiting relationships across categories. The latter does not only let us measure the cross-category similarity but may also have better generalizability.

\begin{figure}
    \centering
    \includegraphics[width=.9\linewidth]{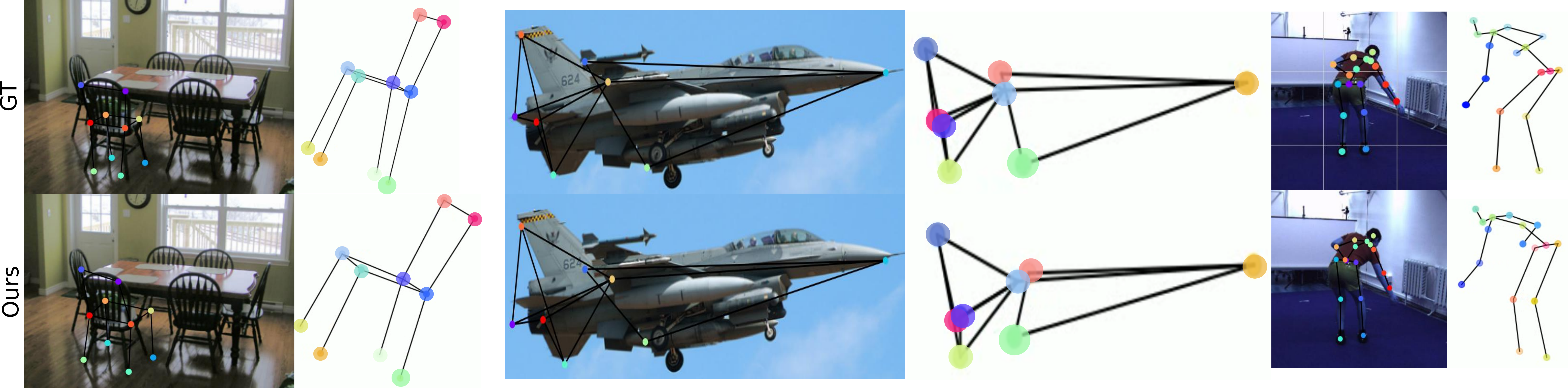}

        \caption{Our method can provide accurate 3D estimations for a wide range of categories directly from a single image. The 2D keypoints are detected and used in conjunction with an image-based feature vector to produce 3D estimates.}
    \label{fig:teaser}
 \vspace{-1em}
\end{figure}

Most existing methods~\cite{DBLP:conf/eccv/ParkLK20,DBLP:conf/iccv/KongL19,wang2021paul,zeng2021pr,novotny2019c3dpo,DBLP:conf/cvpr/ChenTADMSR19} that output pose estimation from images
operate in two stages; 2D keypoint extraction followed by 3D shape and pose estimation. These two stages are often performed independently. We argue that these two stages are dependent and can mutually benefit from each other. Thus, 2D keypoints can be extracted while being suitable for the down-streaming task of 3D reasoning. In particular, we extract the visual context information along with the 2D keypoints from the keypoint extraction network. Later, both visual context and 2D keypoints are provided to the network that lifts 2D keypoints to 3D. Our experiments clearly demonstrate the benefit of visual context information during 3D pose and shape recovery. 

We model the 3D shape using a dictionary learning approach, similar to~\cite{novotny2019c3dpo}, where the shape basis for the union of categories are learned. The instance-wise shape is then recovered with the help of the shape basis coefficients. However, it is known that the size of the shape basis requires careful tuning~\cite{novotny2019c3dpo,DBLP:conf/eccv/ParkLK20}. In a multi-category setting, the latent space is shared by all object categories and each category can have a different optimal shape basis size. Moreover, directly using the shape coefficients results in an over-sensitivity to small perturbations in the input image. We show that both of these problems can be solved through a simple formulation by sparsifying the shape basis by applying a cut-off on the shape coefficients based on a learned threshold vector. This new formulation with a negligible number of additional parameters allows a much simpler network than sparse dictionary-based networks such as~\cite{DBLP:conf/iccv/KongL19}.  

\noindent The major contributions of our work can be summarized as:
\begin{itemize}
\setlength{\itemsep}{0.5pt}
\setlength{\parskip}{0.5pt}
\item End-to-end reconstructing of 3D shape and pose in a multiple category setup, using a single neural network.
\item We propose to use auxiliary image context information to improve the performance.
\item Our method achieves state-of-the-art results in the multi-category setting, with significant improvement.
\end{itemize}

\section{Related Work}

The task of lifting 2D keypoints of deformable objects to 3D from a single image has been mostly studied in the context of NrSfM. In NrSfM, the task is to recover the poses and viewpoints from multiple observations in time of an object~\cite{DBLP:journals/pami/AkhterSKK11}. Significant research has been carried out in NrSfM to improve the performance through sparse dictionary learning~\cite{DBLP:conf/cvpr/KongL16,DBLP:conf/cvpr/ZhouZLDD16}, low-rank constraints~\cite{daubechies2004iterative}, union of local subspaces~\cite{DBLP:conf/cvpr/ZhuHTL14}, diffeomorphism~\cite{DBLP:conf/cvpr/ParasharSF20}, and coarse-to-fine low-rank reconstruction~\cite{DBLP:conf/cvpr/BartoliGCPOS08}. It is possible to use NrSfM frameworks to build category-specific models that can learn to estimate pose and viewpoint from a single image by treating the images of the same category as observations of a single object deformed at different time steps~\cite{DBLP:conf/3dim/KongZKL16,DBLP:conf/iccv/KongL19,DBLP:conf/iccv/ChaLO19}.   
% With the recent surge of deep learning, NrSfM methods experimented using neural networks to obtain 3D locations of given 2D keypoints \cite{DBLP:conf/iccv/ChaLO19, DBLP:conf/iccv/KongL19, DBLP:conf/iccv/NovotnyRGNV19, DBLP:conf/eccv/ParkLK20}. 

Obtaining the 3D structure of an object given only a single image has been studied sparsely. In~\cite{DBLP:conf/eccv/KanazawaTEM18} instance segmentation datasets were utilized to train a model that outputs 3D mesh reconstructions given an image. Correspondences between 2D-3D keypoints were also used to improve results~\cite{DBLP:journals/corr/abs-2106-05662}. While some recent methods can estimate the viewpoint and non-rigid meshes, these methods work on objects with limited diversity, such as faces~\cite{DBLP:conf/ijcai/Wu0V21,DBLP:conf/accv/JenniF20,DBLP:conf/iccvw/SahasrabudheSBG19}. 

The closest line of work to ours involves building a single model for a diverse set of input classes. C3DPO~\cite{DBLP:conf/iccv/NovotnyRGNV19} proposed to learn the factorization of the object deformation and viewpoint change. They propose to enforce the transversal property through a separate canonicalization network that undoes rotation applied on a canonical shape. Park et al. proposed using Procrustean regression~\cite{DBLP:journals/tip/ParkLK18} to determine unique motions and shapes~\cite{DBLP:conf/eccv/ParkLK20}. They also propose an end-to-end method using CNN that can output 3D location of human keypoints from the image. However, their method cannot handle multiple object categories or occluded keypoints. Moreover, it requires temporal information in the form of sequences. Human pose estimation is also tackled in \cite{DBLP:conf/cvpr/ChenTADMSR19}, where the authors propose a cyclic-loss and discriminator. They further boost their results by using temporal information and additional datasets for the training of their GAN. However, their method is limited to human pose estimation. Recently,~\cite{wang2021paul} extended Procrustean formulation with autoencoders and proposed a method that can infer 3D shapes without the need for sequence. However, their method requires a more complex network, two encoders, as well as Procrustean alignment optimization at test-time, which renders the method slow ~\cite{wang2021paul}. All these methods accept 2D keypoints as input rather than images and tackle the problem of obtaining 3D keypoint locations from a single image using a separate keypoint detector, such as a stacked hourglass network~\cite{DBLP:conf/cvpr/ToshevS14}.

\section{Multi-category from a Single View}
We extract 3D structures in the form of 3D keypoints, given only an image of an object category. During training, we only have access to the 2D location of keypoints and the category label. For simplicity, we separate our solution into two parts: category and 2D keypoints extraction from the image and lifting them to 3D. In the following, we will first focus on lifting the given 2D keypoints to 3D. Our approach is formulated in the context of NrSfM, thus we first introduce the preliminaries followed by our approach. We first introduce our lifter network, followed by the end-to-end network introducing our 2D keypoint extractor and the tight coupling with the lifter network. 

\subsection{Preliminaries - NrSfM}

% We extract 3D structures in the form of 3D keypoints, given only an image of some object category. During training, we have access to only 2D location of keypoints, including the category label. For simplicity, we separate our solution into two parts: category and 2D keypoints extraction from the image and lifting them to 3D. In the following, we will first explain our approach for lifting the given 2D keypoints. We formulate the lifting of 2D keypoints problem in the context of NrSfM. 
Let $\mathbf{Y}_i = [\mathbf{y}_{i1}, \ldots , \mathbf{y}_{ik}] \in \mathbb{R}^{2\times k}$ be a stacked matrix representation of $k$ 2D keypoints from the $i^{th}$ view. We represent the structure of the $i^{th}$ view as $\mathbf{X}_i =  \mathbf{\alpha}_i^\intercal \mathbf{S}$, using the shape basis $\mathbf{S}\in\mathbb{R}^{d\times 3k}$ and coefficients $\mathbf{\alpha}_i\in\mathbb{R}^d$. For simplicity, we assume that the keypoints are centered and normalized and that the camera follows an orthographic projection model, represented by $\Pi = [\mathbf{I}_{2\times2}\,\,\, \mathbf{0}]$. Given the camera rotation matrix $\mathbf{R}_i \in \textbf{SO(3)}$, as well as the centered and normalized keypoints, we can write $\mathbf{Y}_i = \Pi \mathbf{R}_i(\mathbb{I}_3 \odot \mathbf{\alpha}_i^\intercal\mathbf{S})$, where the operation $\mathbf{I}_3\odot\mathbf{s}$ reshapes the row vector $\mathbf{s} \in \mathbb{R}^{1 \times 3k}$ to a matrix of the from $\mathbb{R}^{3\times k}$. The recovery of shape and pose by NrSfM given $n$ views can be written as\footnote{We will omit $\mathbb{I}_3$ and the transposition of $\alpha$ throughout the rest of the paper for the ease of notation.},
\vspace{-0.5em}
\begin{equation}
\begin{aligned}
\min_{\mathbf{\alpha}_i,\mathbf{S}, \mathbf{R}_i\in \textbf{SO(3)}} \quad & \sum_{i=1}^n{\mathcal L( \mathbf{Y}_i , \Pi \mathbf{R}_i(\mathbb{I}_3 \odot \mathbf{\alpha}_i^\intercal\mathbf{S}))}.
\end{aligned}
\label{eq:optimizationMain}
\vspace{-0.5em}
\end{equation}
where $\mathcal L (a, b)$ is a norm-based loss of the form $\|a - b\|$.

The above problem is generally ill-posed. Thus, different assumptions regarding $\mathbf{\alpha}_i$ and $\mathbf{S}$ are made in the literature. The most common constraints in this regard are, low-rank~\cite{Dai2012}, finite-basis~\cite{wolf2002projection}, and sparse combination~\cite{zhu2014complex}. In this work, we are interested to solve the problem of~\eqref{eq:optimizationMain} using a learning-based approach. However, we are also interested in a multi-category setting with single view inference.

\begin{figure}
    \centering
    \includegraphics[width=.8\linewidth]{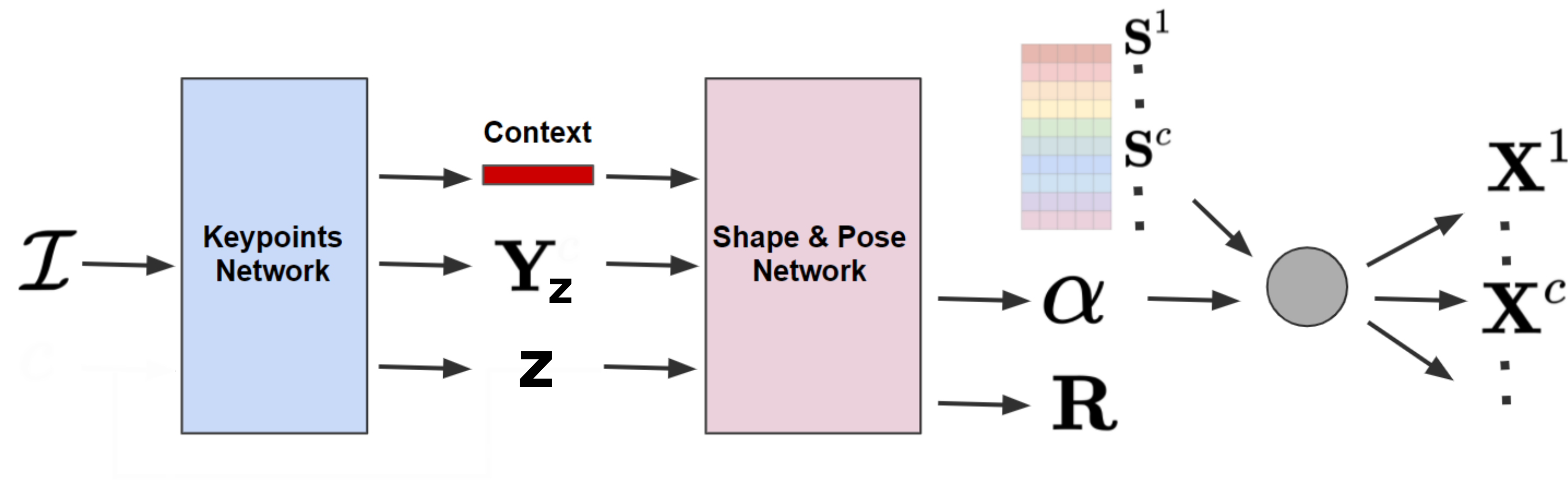}
    
    \vspace{-1.5em}
        \caption{System pipeline. Multi-category from a single view. }
    \label{fig:system}
    \vspace{-1em}
\end{figure}

\subsection{Multi-category Formulation}

In the context of multi-class NrSfM, our method extracts 3D structures of objects from a wide variety of classes. Thus, $(\mathbb{I}_3 \odot \mathbf{\alpha}_i^\intercal\mathbf{S})\in \mathbb{R}^{3\times k}$, should be able to express the 3D structure of objects with different number of keypoints. Let $\textbf{Z}$ represent the set of object categories and $z_i \in \textbf{Z}$ be the category of sample $i$. Let each category $z\in \textbf{Z}$ be represented by $k_z$ keypoints, thus we have a total of $k=\sum_z k_z$ keypoints. To ``access" the correct keypoints we have a subset selection vector $\zeta_z \in \{0,1\}^k$ that indicates which dimensions relate to category $z$. Given these multi-category definitions, we can reformulate \ref{eq:optimizationMain} as
\begin{equation}
\vspace{-0.5em}
\begin{aligned}
\min_{\mathbf{\alpha}_i,\mathbf{S}, \mathbf{R}_i\in \textbf{SO(3)}} \quad & \sum_{i=1}^n{\mathcal L( \mathbf{Y}_i\circ\zeta_{z_i}  , \Pi \mathbf{R}_i( \mathbf{\alpha}_i^\intercal\mathbf{S})\circ\zeta_{z_i})},
\end{aligned}
\label{eq:multiclassMain}
\vspace{-0.5em}
\end{equation}
where $\circ$ is the broadcasted elementwise multiplication.

In the above formulation, $\mathbf{R}_i$ and $\mathbf{\alpha}_i$ are inputs, hence category dependent, while $\mathbf{S}$ is shared among all categories. To formulate the problem as a learning-based approach, let $\mathbf{\alpha}_i$ be the output of a function of input $\mathbf{Y_i}$, i.e. $\alpha(\mathbf{Y_i})$. Let us separate the function $\alpha(.)$ into two composite functions $\alpha(\mathbf{Y_i}) = g(f(\mathbf{Y}))$, with $g(.)$ being an affine function, $g(\mathbf{\upsilon}) = W_g \mathbf{\upsilon} + b_g$ with $\mathbf{\upsilon}\in \mathbb{R}^{F}, W_g\in \mathbb{R}^{D\times F}, b_g\in \mathbb{R}^{D}$. We do not place any restriction on the function $f(.)$ other than taking some observation $\mathbf{Y_i}$ and outputting a vector of dimension $F$. Moreover, let us rewrite $\mathbf{R}_i$ as a function of the input $y$, i.e. $R(\mathbf{y})$. Representing all the parameters with $\theta$, the problem definition becomes
\begin{equation}
    \min_{\theta} \sum_i \mathcal L\big(\mathbf{Y} \circ\zeta_{z_i} , \Pi \big(R(\mathbf{Y})( [W_gf(\mathbf{Y}) + b_g]\mathbf{S}\big)\circ\zeta_{z_i} \big)\big).
    \label{eq:common}
\end{equation}
In the above formulation, shape basis coefficients $\mathbf{\alpha_i}$ are latent codes with latent space basis vectors $W_g$ and a translation term $b_g$, which are shared for all categories. Projecting features of objects from different categories into a shared latent space lets the method extract cross-categorical geometric relationships, as shown in Fig.~\ref{fig:morph}. Moreover, it substantially simplifies computations since we do not require a separate network for each class.

% -Explain how the shape coefficient $alpha$ is shared across the categories.\\ 
% - Introduce the notations and formulation of multi-category.\\ 
% -- Talk about the benefits, both in theory and experiments. 

\subsection{Cut-off Shape Coefficients}

Equation \ref{eq:common} is under-determined unless there are additional constraints imposed on the system. The most common constraint is restricting the dimension $D$ of the shape basis coefficients $\alpha_n$ \cite{DBLP:conf/cvpr/BreglerHB00,DBLP:journals/pami/AkhterSKK11}. However, selecting the optimal cardinality requires careful hyperparameter tuning \cite{novotny2019c3dpo,DBLP:conf/eccv/ParkLK20}. 

Since our method extracts 3D structures of objects from a wide variety of classes, the latent space has to accommodate latent codes from a wide range of inputs. Considering most objects share some common characteristics, using different manifolds for each class results in failure to utilize cross-class information as well as an increase in the complexity of the method. On the other hand, the dimensionality of the optimal manifold is different for each class. Therefore, ideally, we would like to automatically select a manifold for each input in a way that maximizes the performance. Note that the optimal manifold selection depends not only on the object class, but we would like the method to discover cross-class rules for manifold assignment.
 
% Given a sample $Y_n$, let the estimated basis be denoted by $\widehat{\mathbf{S}}$ and the selected basis vectors among $\widehat{\mathbf{S}}$ be denoted by a binary vector $\mathbb{I}_n \in \{0,1\}^{D}$ where $\sum_d \mathbb{I}_n[d] \leq D; \forall n$, with the basis coefficients $\widehat{\alpha_n} \in \mathbb{R}^{D}$. Then, the representation of $Y_n$ is $\widehat{\psi_n} = (\mathbb{I}_n \circ \widehat{\alpha_n})\widehat{\mathbf{S}}$. 

The manifold selection problem can be posed as integer problem, where given a sample $Y_n$, the network selects a subset of the basis vectors $\mathbf{S}$. This can be formalized using a binary selection vector $\mathbb{I}_n \in \{0,1\}^{D}$ where $\sum_d \mathbb{I}_n[d] \leq D; \forall n$. Given the the basis coefficients $\beta_n \in \mathbb{R}^{D}$, the representation of $Y_n$ is $\psi_n = (\mathbb{I}_n \circ \beta_n)\mathbf{S}$. Since this formulation is non-differential, we propose a differentiable alternative which we call cut-off coefficients. The idea is to truncate negative shape coefficients to zero allowing the network a differential way to select basis vectors. To gain back the expressiveness of full range shape coefficients we additionally introduce a bias term, which allows the network to learn basis which are suited for non-negative coefficients. Thus, we arrive at $\hat{\psi}_n = \beta_n \mathbf{S} + b_S$ where $\beta_n \in \mathbb{R}_{\geq 0}^{D}$ and $b_S \in \mathbb{R}^{B}$. Note that the non-negativity constraint can be simply implemented using a ReLU based truncation ($\beta_n = \text{ReLU}(\beta_n')$).

% However, such an approach is difficult to achieve with a neural network since the selection mechanism is binary and thereby non-differentiable. In this work we actually go a different route and use cot-off shape coefficients. The idea is to truncate negative shape coefficients to zero allowing the network a differential way to select basis vectors. To gain back the expressiveness of full range shape coefficients we additionally introduce a bias term, which allows the network to learn basis which are suited for non-negative coefficients. Thus, we arrive at $\psi_n = \beta_n \mathbf{S} + b_S$ where $beta_n \in \mathbf{R}_{+}^{A \times B}$ and $b_S \in \mathbf{R}_{+}^{B}$. 

% Given a sample $Y_n$, let the selected basis vectors among $\mathbf{S}$ be denoted by a binary vector $\mathbb{I}_n \in \{0,1\}^{D}$ where $\sum_d \mathbb{I}_n[d] \leq D; \forall n$, with the basis coefficients $\alpha_n \in \mathbb{R}^{D}$. Then, the representation of $Y_n$ is $\psi_n = (\mathbb{I}_n \circ \alpha_n)\mathbf{S}$. 

Now let us show that this formulation indeed does not affect the expressiveness of the network, even if we would not re-optimize the basis $\mathbf{S}$. Let a sample $Y_n$, be expressed with the basis coefficients $\alpha_n \in \mathbb{R}^{D}$, i.e. $\psi_n = \alpha_n\mathbf{S}$. Let the proposed representation with cut-off coefficients be $\hat{\psi_n} = \beta_n \mathbf{S} + b_S$. Moreover, let us index the dimensions of $D$ dimensional latent space with $d$. Then the latent space coefficient of sample $Y_n$ for the dimension $d$ is $\alpha_{nd}$, and similarly for $\hat{\psi}_n$, it is $\beta_{nd}$. 
% If we limit the coefficients $\beta_n$ to be non-negative, then ReLU function precisely implements $\mathbb{I}_n \circ \beta_n$. Thus, we can re-write $\hat{\psi_n} = \text{ReLU}(\beta_n)\mathbf{S}$. Now, let us use an abuse of notation and from now on refer to $\text{ReLU}(\beta_n)$ as $\beta_n$ with $\beta_n \geq 0$.
The question is, can we find $\beta_n$ and $b_S$ such that $\hat{\psi_n} = \psi_n$ for all $n$. If this holds, we do not lose any expressive power while arriving in a differentiable manifold selection rule. In mathematical terms this is fulfilled if $\exists \beta_{nd}, b_S$ such that $\sum_d \alpha_{nd}\mathbf{S_d} = \sum_d \beta_{nd}\mathbf{S_d} + b_S, \forall n \leq N$. Note that the bias, which is independent of $n$, is fundamental.
% We will show that if we allow for adding a bias term that is independent of $n$, the equation holds. Thus, the equation becomes $\sum_d \alpha_{nd}\mathbf{S_d} \stackrel{?}{=} \sum_d \beta_{nd}\mathbf{S_d} + b_S, \forall n \leq N$. 

Let $m_d = \min_n \alpha_{nd}$ or in other terms, $m_d$ is the minimum coefficient for latent dimension $d$ among all the data samples $n \leq N$. Then, $\alpha_{nd} - m_d \geq 0, \forall n,d$. This implies that we can set $\beta_{nd} = \alpha_{nd} - m_d$ and $b_S = \sum_d m_d\mathbf{S_d}$. To put it all together, $\sum_d \alpha_{nd}\mathbf{S_d} = \sum_d (\alpha_{nd} - m_d)\mathbf{S_d} + m_d\mathbf{S_d}, \forall n \leq N$ which clearly holds with $\beta_{nd} \geq 0, \forall n,d$. It is easy to see that there exists infinitely many choices for $b_S$, and thus $\beta_n$. One can simply set $b_S = \sum_d (m_d - \epsilon_d)\mathbf{S_d}$ and $\beta_{nd} = \alpha_{nd} - (m_d - \epsilon_d)$, where $\epsilon_d \in \mathbf{R}_{+}, \forall d \in D$. Therefore, the final representation for the data sample $Y_n$ is $\psi_n = \beta_n\mathbf{S} + b_S = \text{ReLU}(\beta_n')\mathbf{S} + b_S$ .

Obviously, we do not know $\alpha_n$, thus $\beta_n$. However, just like training a network to output $\alpha_n$, we can train a network to directly output $\beta_n$ and set the bias term $b_S$ as a learnable parameter to be learned from the data during the training process. The above derivation shows that, by using $\psi_n = \beta_n \mathbf{S} + b_S$ we do not lose any expressive power and still represent any sample as accurately as the common formula employed in literature, $\psi_n = \alpha_n \mathbf{S}$.

\begin{figure}
    \centering
    \includegraphics[width=.6\linewidth]{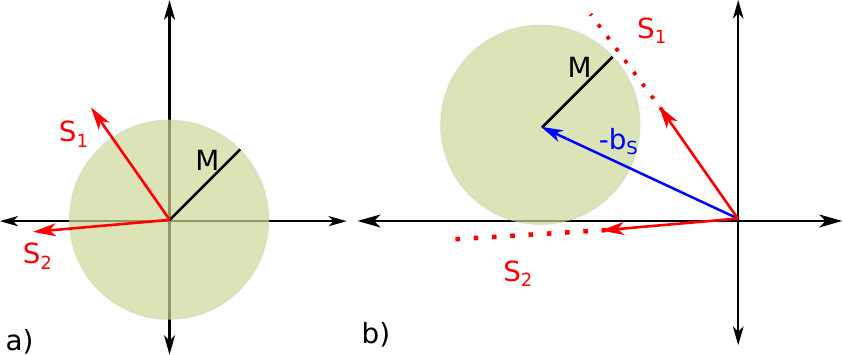}
\vspace{-0.5em}
        \caption{With cut-off coefficients, we translate the latent space by $b_S$, which is learned from the data. }
    \label{fig:bijection}
 \vspace{-0.5em}
\end{figure}

More precisely for the training using our cut-off approach we rewrite \eqref{eq:common}, to,
\begin{equation}
   \mathcal L\big(\mathbf{Y} \circ\zeta , \Pi \big(R(\mathbf{Y})( \text{ReLU}(W_gf(\mathbf{Y}) + b_g)\mathbf{S} + b_S\big)\circ\zeta \big)\big)\,.
    \label{eq:base}
\end{equation}
Thus, we get the discussed advantages, mainly that our method learns to adaptively pick the active basis vectors, thus selecting the dimensionality of the manifold. Furthermore, all parts including the shape basis vectors $\mathbf{S}$, the coefficient generating function $W_gf(.)+b_g$ and the bias $b_S$, are learned from the data. Note that the proposed formulation does not require ISTA iterations \cite{daubechies2004iterative} which is employed in \cite{DBLP:conf/iccv/KongL19} through a specific encoder-decoder architecture. Moreover, the learnable bias term differentiates our method from sparsity iteration and allows for preserving the expressive power of the method. 
%The manifold selection rule is embedded in the cut-off operation applied by ReLU. 

An additional advantage is that the proposed formulation has the property of allowing for sparseness. This in terms encourages the representation of the objects in shape space, i.e. coefficients $\beta$ to be disentangled. Intuitively, as the number of active (non-zero) coefficients increases, more different combinations of the shape basis vectors $\mathbf{S}$ can arrive at the same solution. By allowing to automatically cut off coefficients, the network can learn a small number of shape coefficients to represent changes from one object to another, thus each coefficient can learn to represents a different major variation. 
% Moreover, the cut-off imposed by the ReLU implies that a small change in the coefficients will, likely, not result in any change in the output if the coefficient is inactive. This improves the robustness of the overall method. 
Moreover, sparse representations can promote disentanglement \cite{DBLP:journals/jmlr/GlorotBB11,DBLP:journals/ftml/Bengio09}. In our case, this is especially beneficial since we want the common latent space to encode inter-categorical geometric relationships. Our experiments show that using the proposed formulation leads to a better disentangled representation. 

% - Discuss in the context why are we doing it, what are the benefits, how do we do it by ReLU.

% \subsection{Shape and Pose Loss}
% - Introduces losses of the second block only.

\section{End-to-End Learning from Images}

The image of an object can be used for more than only 2D keypoint extraction. We propose to detect the 2D keypoints from the image and extract a context vector that can be used in conjunction with the 2D keypoints to obtain a better 3D estimation. The detected 2D keypoints and context vector are used by the lifter network, in our end-to-end trainable pipeline.

\begin{figure*}
    \centering
    \includegraphics[width=.9\linewidth]{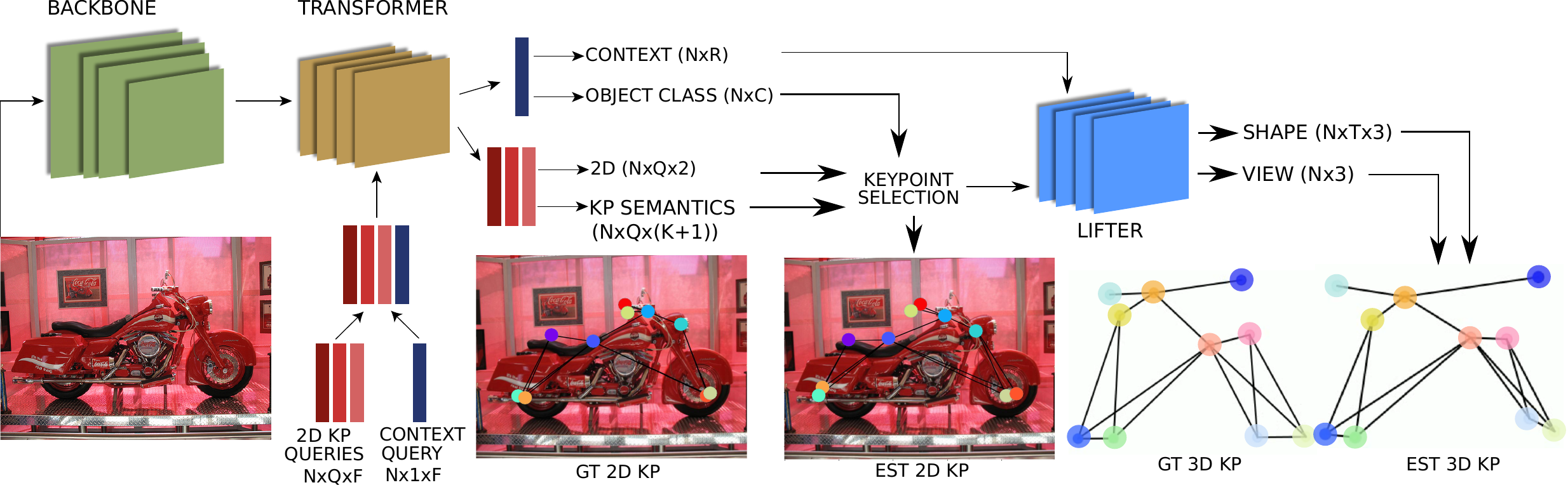}
    \vspace{-1em}
        \caption{Our method uses a transformer architecture to extract 2D keypoints, object class, and a context vector. The detected keypoints and the context vector are processed by a fully connected network to output 3D keypoints and viewpoint from only a single image. The whole end-to-end training is only supervised by 2D keypoint annotations.}
    \label{fig:transformer}
 \vspace{-1em}
\end{figure*}

\subsection{Keypoints from Images}

We require a method that can output the locations of an object category dependent pre-defined set of keypoints. Therefore, the problem at hand naturally extends to object classification. Moreover, in order to fully utilize the image, the keypoint network should produce a context feature representation from the image that can guide the lifter network. Thus, the desired function is $T(\mathcal I) = (\mathbf{Y_\mathbf{z}}, \mathbf{z}, \rho)$ where $\mathcal I$ is the image, $\mathbf{Y_z}$ is the category dependent keypoint locations, $\mathbf{z}$ the category of the object and $\rho$ is the context vector.

\begin{figure}
    \centering
    \includegraphics[width=.8\linewidth]{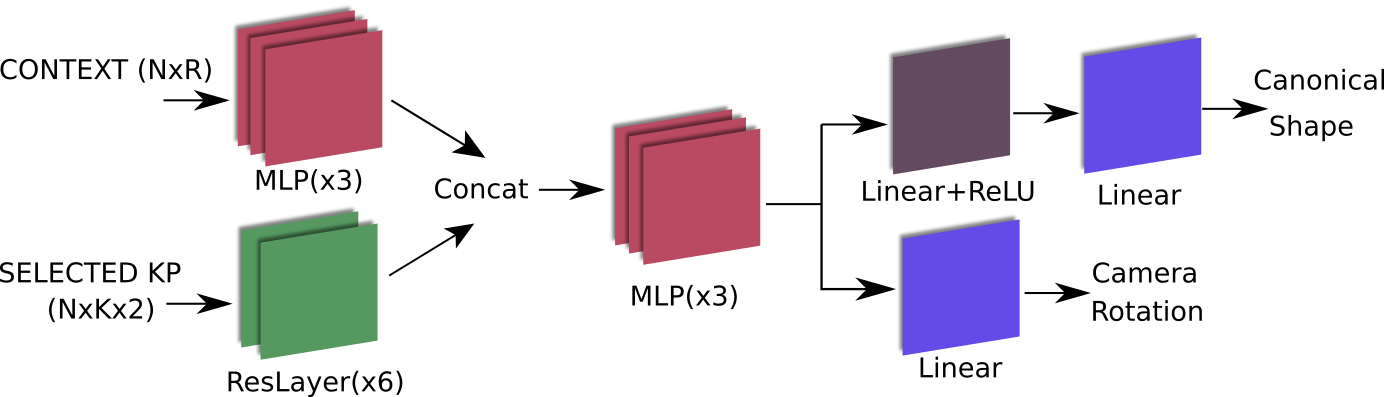}
     \vspace{-1em}
        \caption{Our lifter architecture combines context vector with the estimated keypoints to produce improved pose estimates.}
    \label{fig:lifter}
 \vspace{-1em}
\end{figure}

We propose to use a DETR-based~\cite{DBLP:conf/eccv/CarionMSUKZ20} architecture at the core of function $T$. Thus, the input image is processed by the backbone (Resnet50~\cite{DBLP:conf/cvpr/HeZRS16}), and the resulting feature map is fed to the transformer. The transformer uses two sets of learned query vectors. The first set is related to keypoints, where each query vector $q$ represents a keypoint. To formalize this, let the maximum number of keypoints among all categories be $\text{max}_z k_z $ be $K$. Thus, we can extract 2D normalized locations $\delta \in [0,1]^2$ and the semantics $\omega \in \{0,1\}^K$ of each keypoint by processing the corresponding query vector using two MLPs. The semantics of a keypoint is category dependent and encoded as a one-hot vector. For example, a given entry in $\omega$ can correspond to the front right of a car or the left rear leg of a chair. Entries of $\omega$ with indices larger than $k_z$ are zero. The true semantics are denoted by $\Omega$.

% Our approach is based on DETR \cite{DBLP:conf/eccv/CarionMSUKZ20}. The input image is processed by the backbone (Resnet50 \cite{DBLP:conf/cvpr/HeZRS16}), and the resulting feature map is fed to the transformer. We use a set of learnt queries for the keypoint detection. Moreover, we use a separate learnt query vector to summarize the context. The keypoint query vectors are processed by two different MLPs to obtain the normalized image coordinates and the type of the keypoints. Type of a keypoint encodes the semantic meaning of the keypoint and dependent on the object class. The type vector's size is set to the maximum number of keypoints among all classes, i.e., $\text{max}_b k_b$. With this definition, a given position in the type vector can correspond to front right of a car or left rear leg of a chair.

To help the lifter network estimate the 3D keypoints, the visual context in the image is important, this is especially true in multi-category examples with large in-category variations. Thus, we use a second set of learned query vectors, which gets processed by the transformer together with the keypoint queries. The output of the transformer for the context query is then processed by two MLPs. The first outputs a $N_\rho$ dimensional context vector and the second $\hat{\mathbf{z}}$ the one-hot encoded category probability. The category probability is used in conjunction with the keypoint type estimates to obtain the correct 2D keypoint representation, while the context vector is used by the lifter network together with the 2D keypoint representation, see Fig.~\ref{fig:transformer}.

% In order to fully utilize the image, we need more than only extracting 2D keypoints. The context query vector is processed by two separate MLPs to obtain the class probability of the object in the image and a context vector that encodes additional information. The class estimation is used in conjunction with the keypoint type estimates to obtain the correct 2D keypoint representation while the context vector is used by the lifter network together with the 2D keypoint representation. 
We train the network with two supervision signals. First, we perform direct supervision of the 2D keypoints and the category-specific outputs $\omega$ and $c$, where we use Hungarian matching to select the supervision targets. Second, by training end-to-end, the keypoint extraction network also receives supervision information via the lifter network, which helps to learn the lifting and keypoint regression jointly. It is also the indirect supervision signal that guides the learning of the context $\rho$. This end-to-end connection of the lifter and keypoint extraction network is in sharp contrast to existing papers, which focused on either of the two parts. We will show in our result section that the combination of the two can greatly improve performance.

\subsection{End-to-end Pipeline}

Given the end-to-end joint 2D-3D model, the first step in the training loop is Hungarian matching over the keypoint queries and the GT keypoints. For this, the loss to minimize is given by $\mathcal L_H = \mathcal L_l + \mathcal L_k$ where $\mathcal L_l = ||y - \delta||_1$ and $\mathcal L_k = \mathcal L_{CE}(\Omega, \omega)$. The Hungarian matching output provides the set of query vectors that are one-to-one matched to true keypoints. We reformat selected location estimates $\hat{\delta}$ using the matched semantics $\hat{\omega}$ and the category estimate $\hat{\mathbf{z}}$ into the form given in Eq.~\ref{eq:multiclassMain}. Let this extracted 2D keypoint representation be $\hat{\mathbf{Y}}$ and the true keypoints be $\bar{\mathbf{Y}}$. Adding the category loss of the keypoint network, we arrive at the following set of losses: 
% Given the end-to-end joint 2D-3D model, the first step in the training loop is Hungarian matching over the keypoint queries and the GT keypoints. For this, the loss to minimze is given by $\mathcal L_H = \mathcal L_l + \mathcal L_k$ where $\mathcal L_l = \norm{y - \delta}_1$ and $\mathcal L_k = \mathcal L_{CE}(\Omega, \omega)$. The Hungarian matching output provides the set of query vectors that are one-to-one matched to true keypoints. We reformat selected location estimates $\hat{\delta}$ using the matched semantics $\hat{\omega}$ and the category estimate $\hat{\mathbf{z}}$ into the form given in Eq.~\ref{eq:multiclassMain}. Let this extracted 2D keypoint representation be $\hat{\mathbf{Y}}$ and the true keypoints be $\bar{\mathbf{Y}}$. Adding the category loss of the keypoint network, we arrive at the following set of losses. 
\begin{itemize}
\vspace{-0.5em}
\setlength{\parskip}{0.5pt}
    \item Location loss $\mathcal L_l =||\bar{\mathbf{Y}} - \hat{\delta}||_1$
    \item KP Type loss $\mathcal L_k = \mathcal L_{CE}(\Omega, \hat{\omega})$
    \item Category loss $\mathcal L_b =\mathcal L_{CE}(\bar{\mathbf{z}}, \hat{\mathbf{z}})$
    \item Reprojection loss $\mathcal L_r = \mathcal L(\bar{\mathbf{Y}} \circ\zeta, \Pi R(\hat{\mathbf{Y}}, \rho) f(\hat{\mathbf{Y}}, \rho) \circ\zeta )$
\end{itemize}
where we use Huber loss for $\mathcal L_r$. For the total loss, the different terms are combined using hyperparameters. 

During the evaluation, where we cannot use Hungarian matching, we first get the object category estimate $\hat{\mathbf{z}}$. Then, for each keypoint type defined for that category, we take the location of the most likely proposal and convert it into the form given in Eq.~\ref{eq:multiclassMain} to obtain $\hat{y}$. The combination of $\hat{y}$ and the context vector is processed by the lifter network which outputs the 3D pose and view.

The lifter network is given in Fig.~\ref{fig:lifter}. The architecture is designed to allow for easy pre-training. We first pre-train the transformer to estimate the locations of 2D keypoints and feed the lifter network true 2D locations alongside the context vector. After that we end-to-end train the whole method.

\section{Experiments}

\subsection{Datasets}

We experiment on the \textbf{Synthetic Up3D (S-Up3D)}, \textbf{PASCAL3D+} and \textbf{Human3.6M} datasets. For all datasets, we use the pre-processed versions of \cite{novotny2019c3dpo}. 

% 3 datasets that offer different challenges.

% \noindent\textbf{Synthetic Up3D (S-Up3D).} Unite the People dataset \cite{DBLP:conf/cvpr/Lassner0KBBG17} provides dense human keypoints. The samples originally have 6890 vertices. Following \cite{DBLP:conf/cvpr/Lassner0KBBG17}, we use only 79 vertices. We use the same train/test split as \cite{DBLP:conf/iccv/NovotnyRGNV19}.

% \noindent\textbf{PASCAL3D+} \cite{DBLP:conf/wacv/XiangMS14} provides 2D and 3D keypoints for 12 object classes. The images are from PASCAL VOC and ImageNet datasets and the 3D keypoints are obtained from manually fitting CAD models. We follow the canonical train/test split. 

% \noindent\textbf{Humans3.6M} \cite{DBLP:journals/pami/IonescuPOS14} is one of the largest human pose datasets. While it provides 3D locations of 32 keypoints, following the previous works, we only consider 17 joints.

\begin{figure}
    \centering
    \includegraphics[width=.9\linewidth]{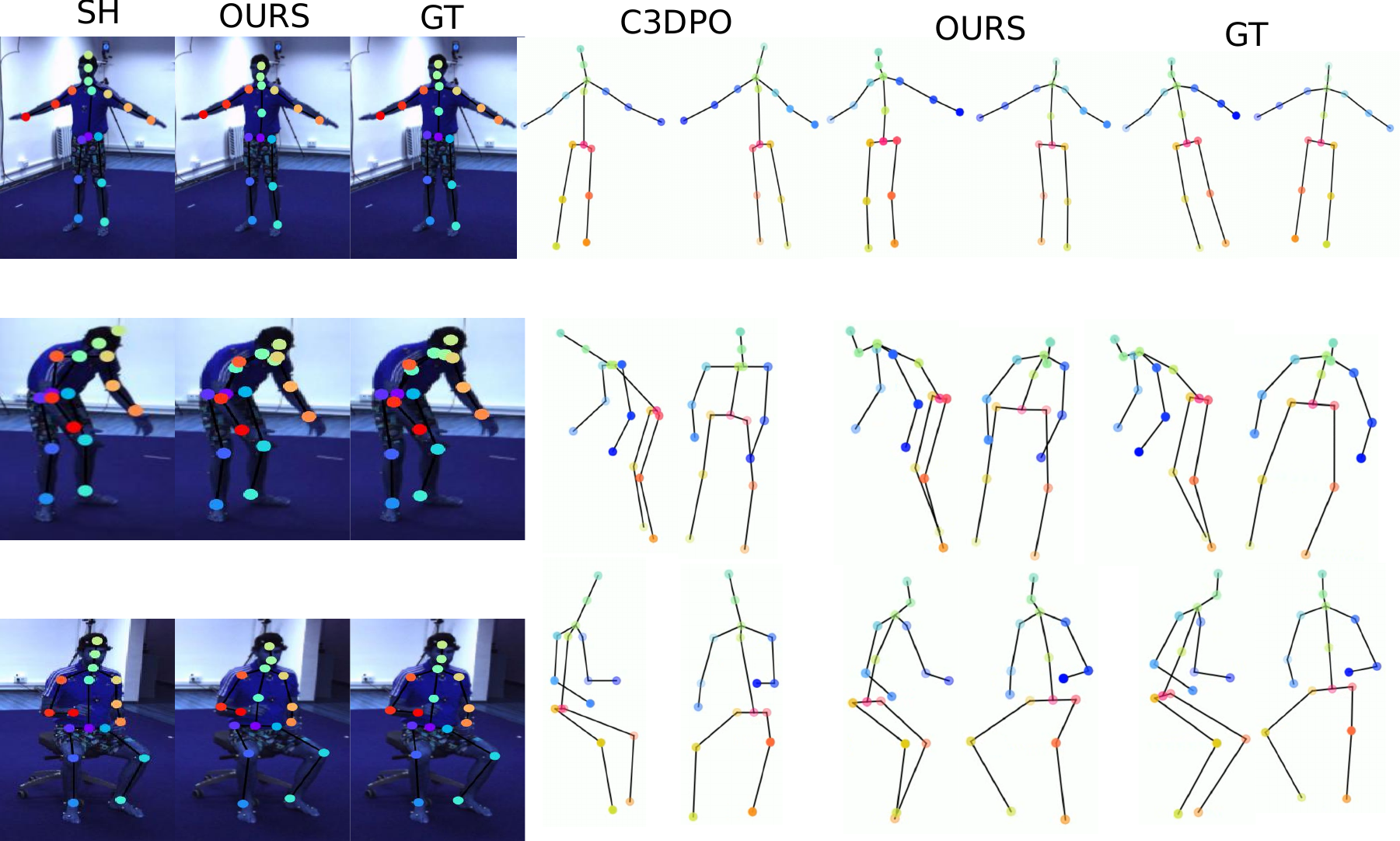}

        \caption{Visual results on the Human3.6M dataset. 2D keypoint estimates of ours, HRNet (SH) from \cite{novotny2019c3dpo} an GT, plus 3D structures from 2 angles for C3DPO, ours and GT. The images in the figure are cropped around the humans for visualization purposes. 2D keypoint results clearly demonstrate the superiority of our method over the commonly used keypoint detector. The performance of our method is also visible in 3D results.}
    \label{fig:human_results}
 \vspace{-1em}
\end{figure}

\subsection{Baselines}

There are only a few NrSfM methods that can handle a setting as diverse as our method. We compare against C3DPO~\cite{DBLP:conf/iccv/NovotnyRGNV19} and PAUL~\cite{wang2021paul} in all datasets since they can produce accurate estimates in a wide range of datasets and settings. We also report results of EMSfM~\cite{DBLP:journals/pami/TorresaniHB08} and GbNrSfM~\cite{DBLP:conf/nips/FragkiadakiSAM14} on the S-Up3D and PASCAL3D datasets. We compare against~\cite{DBLP:conf/eccv/ParkLK20,DBLP:journals/corr/abs-1803-08244,DBLP:conf/cvpr/ChenTADMSR19} only in Human3.6M dataset since they cannot handle occlusions or multiple object categories. We refer to the end-to-end method of~\cite{DBLP:conf/eccv/ParkLK20} as Proc-CNN. Note that, obtaining an Orthographic-N-point (OnP)~\cite{steger2018algorithms} solution requires an optimization at test time which renders methods that depend on OnP~\cite{park2017procrustean,wang2021paul} slower than feed-forward methods such as ours. Also, \cite{DBLP:conf/cvpr/ChenTADMSR19}, which we refer to as Geo, uses extra datasets and temporal information. In Pascal3D, we also compare against CMR~\cite{DBLP:conf/eccv/KanazawaTEM18}. %We report results of our lifter network without (Ours-base) and with cycle-supervision (Ours).

To validate the end-to-end pipeline, we report results with three settings: 1) Lifter and transformer are trained separately without context vector (Ours/TR); 2) Lifter and transformer are trained end-to-end without context vector (Ours w/o Context); 3) The proposed end-to-end training with context vector (Ours). Moreover, we also experiment with using the stacked hourglass network~\cite{DBLP:conf/cvpr/ToshevS14} to extract 2D keypoints on the Pascal3D and Human3.6M datasets.

\subsection{Evaluation protocol}

Following~\cite{DBLP:conf/iccv/NovotnyRGNV19}, we report absolute mean per joint position error $\textbf{MPJPE}(X,Y)=\sum_{k=1}^K||X_k - Y_k||/K$ as well as $\textbf{Stress}(X,Y)=\sum_{i<j}| ||X_i - X_j|| - ||Y_i - Y_j|| |_1/(K(K-1))$, where we center both the estimates and ground truth at zero mean. To calculate MPJPE, we flip the depth dimension and calculate MPJPE for both cases, keeping the best as our result. This is done to resolve the depth ambiguity problem where the network can assign the depth negative or positive direction. For all datasets, we follow the same canonical train/test split as~\cite{DBLP:conf/iccv/NovotnyRGNV19}.

\subsection{Implementation}
For experiments with GT keypoints, where only our lifter network is used, we only use visible keypoints as inputs. We mask out the invisible points and concatenate the visibility mask along with the input keypoints (following~\cite{DBLP:conf/iccv/NovotnyRGNV19}). Our joint end-to-end model takes an image and outputs the 3D location of all the keypoints. In order to provide a fair comparison, we use an architecture with a similar parameter count as~\cite{DBLP:conf/iccv/NovotnyRGNV19} for our lifter network. Moreover, we follow the same data pre-processing as~\cite{DBLP:conf/iccv/NovotnyRGNV19}. Our implementation is in Pytorch and our end-to-end joint method processes one frame in 0.024s on an RTX2080 GPU.

% We first pre-train the lifter network without cyclic supervision. In order to apply cyclic estimation, we update the batch norm running variables in the first forward pass and freeze them while calculating the cycle estimates. This allows us to circumnavigate the instability issues of batch norm when used in recursions.

\section{Results}

In order to show the performance of individual components, we separate the results into two parts: (i) using GT 2D keypoints and (ii) estimating the keypoints directly from the image and producing the 3D pose with these estimates. 

\begin{figure}
    \centering
    \includegraphics[width=.9\linewidth]{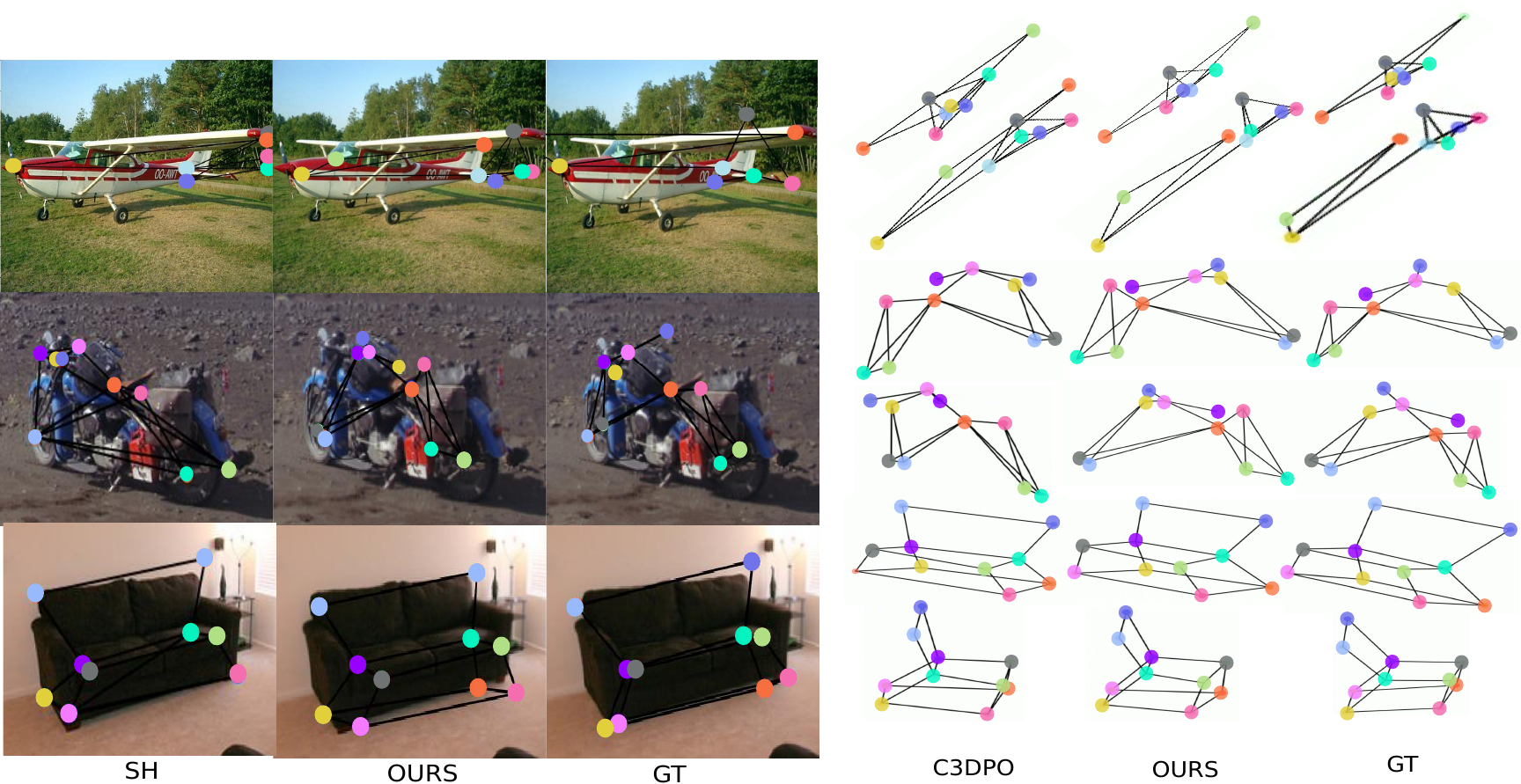}
\vspace{-1em}
        \caption{Visual results on the Pascal3D dataset. 2D keypoint estimates of ours, HRNet (SH) from \cite{novotny2019c3dpo} an GT, plus 3D structures from 2 angles for C3DPO, ours and GT. The images in the figure are cropped around the humans for visualization purposes. Our network produces better 2D keypoint estimations even when the points are occluded. }
    \label{fig:pascal_results}
%  \vspace{-1.5em}
\end{figure}

\begin{table}[h]
% \scriptsize{
\begin{center}

\tabcolsep=0.08cm
\begin{tabular}{ |c|c|c|c|c|c|c| }
\hline
 & \multicolumn{2}{c}{Pascal3D} & \multicolumn{2}{|c|}{Human3.6M}& \multicolumn{2}{|c|}{S-Up3D}\\
\hline
\textbf{Method}& MPJPE & Stress & MPJPE & Stress & MPJPE & Stress\\
\hline
\hline
Geo-SH \textsuperscript{\ddag}* \cite{DBLP:conf/cvpr/ChenTADMSR19}& - & - & 51 & -  & -& -   \\
\hline
EM-SfM \cite{DBLP:journals/pami/TorresaniHB08} & 131.0 & 116.8 & - & -  & 0.107 & 0.061\\

GbNrSfM \cite{DBLP:conf/nips/FragkiadakiSAM14} & 184.6 & 111.3 & - & - & 0.093 & 0.062 \\

PoseGAN  \cite{DBLP:journals/corr/abs-1803-08244}& - & - & 130.9 & 51.8 & - & - \\

Proc \textdagger*\cite{DBLP:conf/eccv/ParkLK20}& - & - & \textbf{86.4} & - & - & -  \\

PAUL \textdagger \cite{wang2021paul}& 30.9 & - & 88.3 & - & 0.058 & -   \\

\hline
C3DPO-base  & 53.5 & 46.8 & 135.2 & 56.9  & 0.160  & 0.105  \\
C3DPO \cite{novotny2019c3dpo}&   38.0 & 32.6 & 101.8 & 43.5 &  0.068 & 0.040  \\

\hline

Ours & \textbf{29.5} &  \textbf{26.6} & 92.8 & \textbf{42.6} & \textbf{0.057} & \textbf{0.035}\\

\hline
\end{tabular}

\end{center}
% }
% \vspace{-1em}
\caption{ Results on Pascal3D, Human3.6M and S-Up3D datasets with GT keypoints. \textdagger: Uses test time optimization. *: Rrequires temporal sequences for training. \ddag Uses additional datasets for training.}
% \vspace{-1em}
\label{tab:gt}
\end{table}
\subsection{Results with GT Keypoints}

We present our results for the lifter network when the GT keypoints are used in Table~\ref{tab:gt}. Our method outperforms all methods that do not perform test-time optimization apart from \cite{DBLP:conf/cvpr/ChenTADMSR19}, which uses additional datasets and temporal information. Our method outperforms all methods in the Pascal3D dataset where our method's multi-class focus is shown best. We also outperform all other methods in the S-Up3D dataset. Comparing C3DPO-base and Ours, the boost the cut-off coefficients provide can be seen. Our method is only slightly worse than the Procrustean network~\cite{park2017procrustean} in the Human3.6M dataset although they use sequences for training and test-time Procrustean optimization. It can be seen that our method produces the best overall results while being applicable in all datasets.

\subsection{Results with Estimated Keypoints}

The results with estimated keypoints, i.e. direct pose estimation from the image, are given in Table~\ref{tab:joint}. In both datasets our method significantly outperforms the competitors. Moreover, we see that the performance boost mainly comes from the proposed joint training and the context vector. Especially test-time Procrustean optimization methods, even when competitive with GT keypoints, suffer considerably using estimated keypoints, visible in the Human3.6M results. For the Pascal dataset, neither PAUL~\cite{wang2021paul} nor Procrustean network~\cite{park2017procrustean} even report numbers. The boost that the context vector provides is rather significant. Note that the context vector also allows for efficient indirect supervision to the transformer through the end task of 3D pose estimation. Our method provides the best overall results in different datasets. This is the result of the proposed flexible end-to-end framework that can be readily applied to any dataset.
% cannot work with imperfect keypoint estimates since their outputs are obtained by solving for the optimal pose on a loss function directly applied to noisy 2D keypoints. It can be seen that neither PAUL\cite{wang2021paul} nor Procrustean network\cite{park2017procrustean} report number for the Pascal dataset with estimated keypoints.   

\begin{table}[h]
% \scriptsize{
\begin{center}

\tabcolsep=0.08cm
\begin{tabular}{ |c|c|c|c|c| }
\hline
 & \multicolumn{2}{c}{Pascal3D} & \multicolumn{2}{|c|}{Human3.6M}\\
\hline
\textbf{Method}& MPJPE & Stress & MPJPE & Stress\\
\hline
Geo-SH \textsuperscript{\ddag}* \cite{DBLP:conf/cvpr/ChenTADMSR19}& - & - & 68 & -     \\
\hline
CMR/SH   \cite{DBLP:conf/eccv/KanazawaTEM18}& 74.4 & 53.7 & - & -  \\
C3DPO/SH \cite{novotny2019c3dpo} & 57.4 & 41.4 & 145.0 & 84.7  \\
Proc-SH \textsuperscript{\textdagger}* \cite{DBLP:conf/eccv/ParkLK20}& - & - & 124.5 & -    \\
Proc-CNN \textsuperscript{\textdagger}* \cite{DBLP:conf/eccv/ParkLK20}& - & - & 108.9 & -    \\
PAUL-SH \textsuperscript{\textdagger} \cite{wang2021paul}& - & - & 132.5 & -     \\

% CSF2-CNN \cite{DBLP:conf/cvpr/GotardoM11}& - & - & 130.6 & -    \\
% SPM-CNN \cite{DBLP:journals/ijcv/DaiLH14}& - & - & 114.4 & -     \\
% Proc-CNN \cite{DBLP:conf/eccv/ParkLK20}& - & - & 108.9 & -     \\

Ours/SH   & 56.1 &   39.0 & 140.7 & 80.9 \\
Ours/TR   & 61.3 & 47.9 & 114.0 & 58.8 \\
Ours w/o Cont   & 57.6 & 42.9 & 113.8 & 56.7 \\
\textbf{Ours}  & \textbf{46.6} & \textbf{33.6} & \textbf{107.7} & \textbf{55.4} \\
\hline
\end{tabular}

\end{center}
% }
% \vspace{-1em}
\caption{ Results on Pascal3D and Human3.6M datasets. \textdagger: Uses test time optimization. *: Requires temporal sequences for training. \ddag Uses additional datasets for training. }
% \vspace{-1em}
\label{tab:joint}
\end{table}

\subsection{Disentanglement}
In order to evaluate the effect of the proposed cut-off weights on the latent space, we measure the mutual coherence of the latent space basis vectors $W_g$. The linear combinations of these vectors create the latent code that is then decoded into the 3D keypoints via $\mathbf{S}$. Thus, the mutual coherence of the latent basis vectors provides a measure of disentanglement. Table~\ref{tab:mc} shows that sparse cut-off coefficients encourage latent basis vectors to be less correlated. 

\begin{table}[h]
% \scriptsize{
\begin{center}

\tabcolsep=0.08cm
\begin{tabular}{ |c|c|c|c| }
\hline
\textbf{Method} & S-Up-3D & Pascal3D & Human3.6M\\

\hline
\hline

Standard (no cut-off) &  0.89  & 0.84 & 0.44   \\
Ours (cut-off) & 0.36  & 0.38  & 0.24  \\
\hline
\end{tabular}

\end{center}
% }
\vspace{-1em}
\caption{ Mutual coherence of the latent space basis vectors $W_g$ with respect to the proposed cut-off formulation on all datasets.}
% \vspace{-2em}
\label{tab:mc}
\end{table}

\begin{figure}
    \centering
    \includegraphics[width=.9\linewidth]{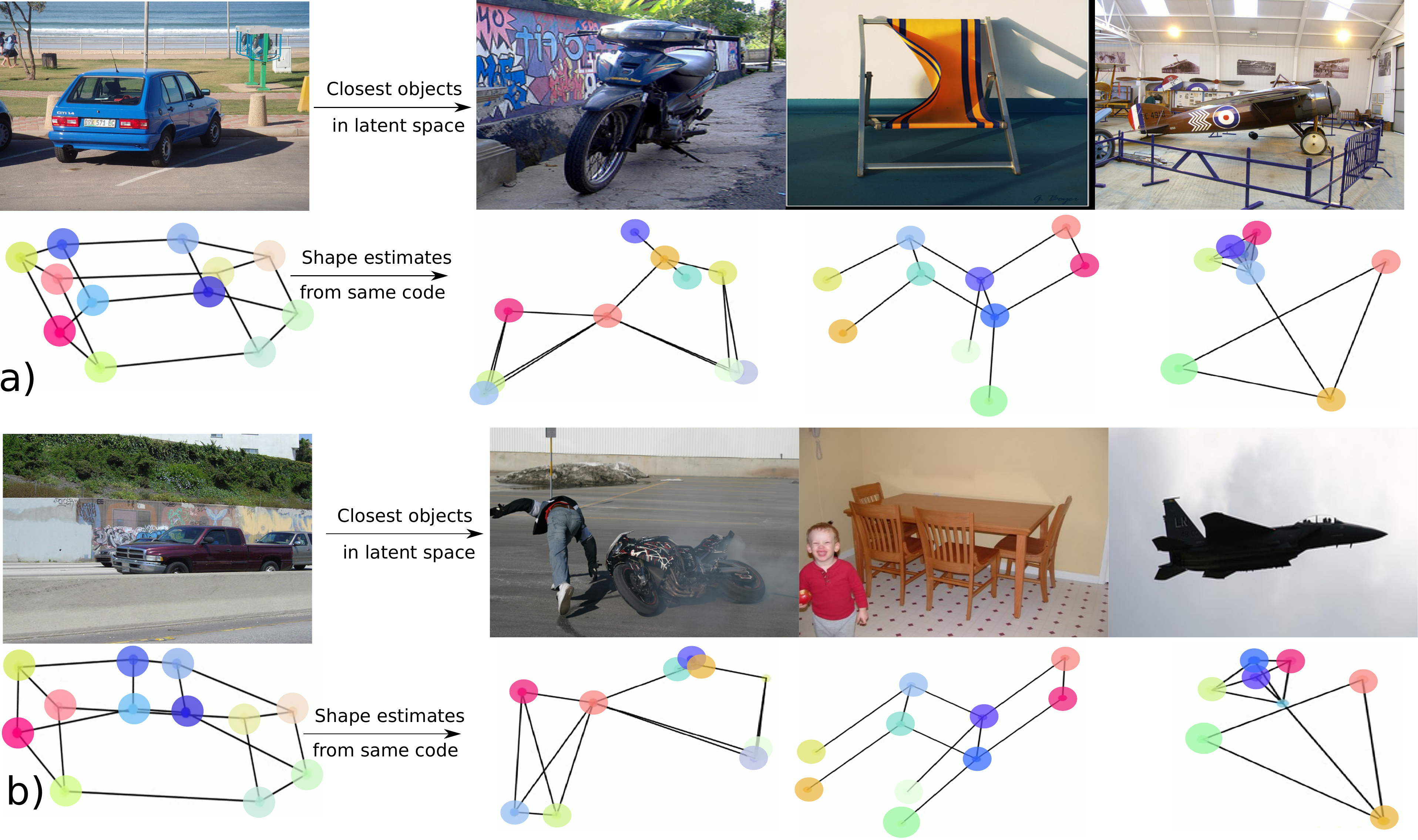}
 \vspace{-1em}
        \caption{Two car samples from the validation set in (a) and (b). The images are the per-category closest samples in the latent space. The 3D reconstructions are from the latent code of the corresponding car mapped into different categories. Thus 3D shapes are \underline{not} of the images but rather decoding of the latent codes of the cars. Elongation (height/width ratio) is translated across categories. }
    \label{fig:morph}
%  \vspace{-1.5em}
\end{figure}

We further demonstrate the properties of the latent space of the proposed formulation in Fig.~\ref{fig:morph}. It can be seen that the latent space codes translate across object categories and encode geometric properties such as elongation. Moreover, the estimated 3D shapes of different categories from the same latent code match closely with the images that reside close to each other in the latent space. This proves our method's capability of successfully handling different classes.
 
% \newpage
\section{Conclusion}
We study the problem of estimating 3D pose and shape from a single image for objects of multiple categories, in an end-to-end manner. Our learning framework relies only on 2D annotations of keypoints for supervision, and exploits the relationships between keypoints within and across categories. The proposed end-to-end learning process offers a structured and unified approach for the image-to-3D problem. Our extensive experiments show that end-to-end training improves the performance substantially. Moreover, the use of contextual information in estimating the 3D pose boosts the performance significantly. Our method is the first of its kind, providing a framework that can be applied to any dataset. We also outperform all the compared methods in estimating 3D shape and pose directly from images, on three benchmark datasets.  \\ 

\clearpage
\bibliographystyle{splncs04}
\bibliography{egbib}
\end{document}